\documentclass[11pt]{article}

\usepackage{arxiv}
\renewcommand{\shorttitle}{The Exceedance Design Effect}
\usepackage[font=small,labelfont=bf,margin=1.5em]{caption}

\usepackage[utf8]{inputenc}
\usepackage[T1]{fontenc}

\usepackage{pifont}
\DeclareUnicodeCharacter{0394}{\ensuremath{\Delta}}
\DeclareUnicodeCharacter{03A3}{\ensuremath{\Sigma}}
\DeclareUnicodeCharacter{03B4}{\ensuremath{\delta}}
\DeclareUnicodeCharacter{03BB}{\ensuremath{\lambda}}
\DeclareUnicodeCharacter{03C1}{\ensuremath{\rho}}
\DeclareUnicodeCharacter{03C3}{\ensuremath{\sigma}}
\DeclareUnicodeCharacter{03C4}{\ensuremath{\tau}}
\DeclareUnicodeCharacter{2194}{\ensuremath{\leftrightarrow}}
\DeclareUnicodeCharacter{2208}{\ensuremath{\in}}
\DeclareUnicodeCharacter{2212}{\ensuremath{-}}
\DeclareUnicodeCharacter{222A}{\ensuremath{\cup}}
\DeclareUnicodeCharacter{2248}{\ensuremath{\approx}}
\DeclareUnicodeCharacter{2260}{\ensuremath{\neq}}
\DeclareUnicodeCharacter{2261}{\ensuremath{\equiv}}
\DeclareUnicodeCharacter{2264}{\ensuremath{\leq}}
\DeclareUnicodeCharacter{2265}{\ensuremath{\geq}}
\DeclareUnicodeCharacter{2308}{\ensuremath{\lceil}}
\DeclareUnicodeCharacter{2309}{\ensuremath{\rceil}}
\DeclareUnicodeCharacter{2713}{\ding{51}}
\DeclareUnicodeCharacter{2717}{\ding{55}}
\usepackage{hyperref}
\usepackage{url}
\usepackage{booktabs}
\usepackage{amsfonts}
\usepackage{nicefrac}
\usepackage{microtype}
\usepackage{lipsum}
\usepackage{graphicx}
\usepackage{amsmath}
\usepackage{amssymb}
\usepackage{natbib}
\usepackage{longtable}
\usepackage{needspace}
\makeatletter
\g@addto@macro\@verbatim{\footnotesize}
\makeatother

\usepackage{array}
\usepackage{multirow}
\usepackage{calc}

\providecommand{\tightlist}{%
  \setlength{\itemsep}{0pt}\setlength{\parskip}{0pt}}

\providecommand{\pandocbounded}[1]{%
  \resizebox{\ifdim\width>\linewidth\linewidth\else\width\fi}{!}{#1}}

\title{The Exceedance Design Effect:\\ Effective Sample Size for Thresholds under Clustering}

\author{
  Adam Noonan\thanks{ORCID: \href{https://orcid.org/0009-0000-7506-6395}{0009-0000-7506-6395}. First posted 26 July 2026; also \href{https://arxiv.org/abs/2608.21262}{arXiv:2608.21262}. Archival version and verification code: \href{https://doi.org/10.5281/zenodo.21595640}{doi:10.5281/zenodo.21595640}.} \\
  Independent Researcher \\
  \texttt{adamnoonan@utexas.edu}
}

\date{1 October 2026}

\begin{document}

\maketitle

\section*{Abstract}\label{abstract}
\addcontentsline{toc}{section}{Abstract}

Suppose we want a cutoff that 90\% of a population falls below. We
estimate it from a sample. Another sample gives a different cutoff,
which changes the fraction of the population below it. How much does
that fraction vary when observations come in groups, such as pupils in
classrooms? We study this question for independent groups whose
observations share a continuous distribution. We prove a large-sample
formula for the variance of the population fraction below the cutoff.
For groups of size \(m\), clustering multiplies this variance by
\(1+(m-1)\rho_I(p)\). Here \(p\) is the target fraction. The correlation
\(\rho_I(p)\) measures whether observations in the same group fall on
the same side of the population cutoff. It can differ from correlation
between the scores themselves, and can change with the target fraction,
so a dataset has no single effective sample size. Our result connects
classical methods for clustered quantiles to coverage: the population
fraction below an estimated cutoff. We give a direct proof, a
counterexample to using score correlation, and an extension to unequal
group sizes.

\section{Choosing a cutoff from a
sample}\label{choosing-a-cutoff-from-a-sample}

\subsection{A classroom example}\label{a-classroom-example}

Suppose we want a score that 90\% of pupils fall below. We cannot test
everyone, so we sample some classrooms and test their pupils. We put the
scores in order and choose a cutoff near the top: about 90\% of the
sample falls below it.

Now imagine testing pupils from other classrooms in the same population.
The fraction below our cutoff will usually differ from 90\%. That is
because we chose the cutoff from a sample. A different sample of
classrooms would give us a different cutoff.

Classmates may have similar scores because they share a teacher or a
school. Testing more classmates can therefore tell us less than testing
the same number of pupils from separate classrooms. But similarity in
scores is not quite the question we need to ask. What matters is whether
classmates tend to fall on the same side of the cutoff.

This paper measures how much the fraction below the cutoff varies across
samples. We call that fraction \emph{coverage}. We treat classrooms as
independent groups, called \emph{clusters}. The pupils used to assess
coverage come from other classrooms, sampled independently from the same
population.

\Needspace{7\baselineskip}

\subsection{The result}\label{the-result}

For \(n=bm\) observations from \(b\) independent clusters of size \(m\),
the first-order coverage variance is

\[\frac{p(1-p)}{n}\,[1+(m-1)\rho_I(p)].\]

Here \(p\) is the target coverage and \(\rho_I(p)\) is the
within-cluster indicator correlation at that level. The factor in
brackets is the classical design effect applied to a threshold
indicator. Writing \(n_{\mathrm{eff}}=n/[1+(m-1)\rho_I(p)]\) expresses
its consequence for coverage precision when the factor is positive. This
is a first-order variance equivalence, not an exact replacement for the
sample size in a coverage guarantee.

A dataset does not have one effective sample size. How much information
it contains depends on the question you ask. In the document experiment
of Section 5.3, the same 1,000 rows carried about 217 independent
observations' worth of information at the median. At the 95th
percentile, they carried about 621. Nothing about the dataset changed.
We asked it a different question. That is why a single quality-adjusted
sample count can mislead. The number of rows is a property of the
dataset. The effective sample size belongs to the analysis.

Classical survey methods already use indicator variation to obtain
quantile inference \citep{woodruff1952,franciscofuller1991}.
Clustered-quantile methods also use it in online experimentation
\citep{deng2018,yao2024}. We show directly how clustering affects
coverage under the assumptions in Section 2. We also show why score
correlation can give the wrong answer, and how the answer changes with
the target fraction. The proof uses counts on the probability scale and
requires no positive marginal density. Section 7 compares the result
with its closest predecessors.

Section 2 states and proves the coverage law. Section 3 shows why score
correlation cannot replace indicator correlation. Section 4 treats
operating levels and unequal cluster sizes. Section 5 gives two
empirical illustrations: a released proxy score, and a specified score
on a document corpus. Section 6 explains how to use the result. Section
8 identifies the remaining questions.

\subsection{Application to conformal
prediction}\label{application-to-conformal-prediction}

Split conformal calibration sets the cutoff at rank
\(k=\lceil(n+1)(1-\alpha)\rceil\) among the calibration scores. If
\(k=n+1\), it uses an infinite cutoff. Under exchangeability of all
calibration and test scores, marginal coverage is at least \(1-\alpha\).
Without ties and with \(k\le n\), the uniform-rank argument gives
coverage \(k/(n+1)\).

That statement averages over reference samples. Under independent
continuous sampling, coverage conditional on the reference sample has
distribution \(\mathrm{Beta}(k,n+1-k)\) \citep{vovk2012,bianbarber2022}.
Its variability describes how coverage differs between fitted cutoffs.
This is distinct from the sampling noise of a proportion measured on a
finite test set.

Independent clusters with dependent members do not generally make the
pooled calibration and test scores exchangeable. Consequently, neither
the exact rank guarantee nor the independent-sample Beta law follows
from our assumptions. Theorem 1 instead describes the first-order
coverage fluctuations as the number of clusters grows. The mean can also
change, and Section 2.4 shows that the indicator correlation alone does
not determine it.

\section{The coverage law}\label{the-coverage-law}

\subsection{Setup}\label{setup}

Let \(S_{j,r}\) be observation \(r\) in cluster \(j\). There are \(b\)
clusters of fixed size \(m\ge2\), giving \(n=bm\) observations. Fix a
target level \(p_0\in(0,1)\) and assume:

\begin{itemize}
\tightlist
\item
  \textbf{(A1)} Clusters are independent and identically distributed.
  Observations are exchangeable within each cluster.
\item
  \textbf{(A2)} Each observation has the same continuous marginal CDF
  \(F\).
\item
  \textbf{(A3)} The test observation is independent of the reference
  sample and has CDF \(F\).
\end{itemize}

The within-cluster exchangeability assumption can be relaxed, as Section
2.3 explains. Continuity does not require a positive density, but it
excludes marginal atoms. Randomization can produce continuous scores; it
also changes the procedure being evaluated (Appendix B.1).

Set \(k=\lceil(n+1)p_0\rceil\) and \(p_n=k/(n+1)\). Then
\(0\le p_n-p_0<1/(n+1)\). For \(k\le n\), the cutoff is
\(\hat q=S_{(k)}\). For \(k=n+1\), use an infinite cutoff; this case
disappears eventually as \(b\) grows. The population fraction below the
fitted cutoff is

\[C=\mathbb P(S_{\mathrm{test}}\le\hat q\mid\text{reference sample})=F(\hat q).\]

We call \(C\) the \emph{calibration-conditional coverage}. It varies
across reference samples even when the test population stays fixed.

For \(p\in(0,1)\), write \(q_p=F^{-1}(p)\) and define

\[\delta(p)=\mathbb P(S_{1,1}\le q_p,S_{1,2}\le q_p),\qquad
\rho_I(p)=\frac{\delta(p)-p^2}{p(1-p)}.\]

Thus \(\rho_I(p)\) is the within-cluster correlation of
\(I_{j,r}(p)=\mathbb1\{S_{j,r}\le q_p\}\). Complementing both indicators
gives the same correlation for exceedances above the cutoff. We call

\[D(p)=1+(m-1)\rho_I(p)\]

the \emph{exceedance design effect}. It is the classical design effect
applied to the indicator \citep{kish1965,katz1993}.

\subsection{Statement and proof}\label{statement-and-proof}

\begin{quote}
\textbf{Theorem 1.} Under (A1)--(A3), as \(b\to\infty\) with \(m\)
fixed,
\[\sqrt n(C-p_n)\xrightarrow{d}\mathcal N\!\left(0,p_0(1-p_0)D(p_0)\right).\]
Moreover, \(n\operatorname{Var}(C)\to p_0(1-p_0)D(p_0)\).
\end{quote}

For \(D(p)>0\), define \(n_{\mathrm{eff}}(p)=n/D(p)\). The corresponding
first-order variance is \(p(1-p)/n_{\mathrm{eff}}(p)\). At \(D(p_0)=0\),
the limit is degenerate and \(n\operatorname{Var}(C)\to0\). An infinite
first-order effective size in that case does not mean zero finite-sample
uncertainty.

The proof has two steps: apply a CLT to independent cluster counts, then
invert the count event. The final tail bound justifies convergence of
moments.

\textbf{Proof.} Put \(U_{j,r}=F(S_{j,r})\). Assumption (A2) makes each
marginal uniform, and monotonicity gives \(C=U_{(k)}\) exactly. Define
\(X_j(t)=\sum_{r=1}^m\mathbb 1\{U_{j,r}\le t\}\) and
\(N_b(t)=\sum_{j=1}^bX_j(t)\). The counts are independent across
clusters, have mean \(mt\), and satisfy

\[\operatorname{Var}X_j(t)=mt(1-t)D(t),\qquad
D(t)=1+(m-1)\rho_I(t).\]

The pair-copula diagonal is Lipschitz, so \(D\) is continuous on
interior levels. For \(v=p_0(1-p_0)D(p_0)>0\), the i.i.d. CLT gives
\([N_b(p_0)-np_0]/\sqrt n\Rightarrow\mathcal N(0,v)\). For an interval
of probability \(h\), the count increment \(L_j\) satisfies
\(0\le L_j\le m\) and \(\mathbb E L_j=mh\), hence
\(\operatorname{Var}L_j\le m^2h\). Thus replacing \(p_0\) by any
deterministic \(t_b\to p_0\) changes the centered, normalized count by a
term with variance at most \(m|t_b-p_0|\to0\).

Take \(t_b=p_n+x/\sqrt n\). The exact order-statistic identity is

\[\mathbb P\{\sqrt n(C-p_n)\le x\}
=\mathbb P\{N_b(t_b)\ge k\}.\]

Since \((k-nt_b)/\sqrt n=p_n/\sqrt n-x\to-x\), the CLT and the increment
bound give the claimed limit. If \(v=0\), the normalized count has
variance tending to zero. Chebyshev's inequality at each fixed positive
and negative \(x\) gives convergence to zero in probability.

For completeness, bounded cluster counts also control moments.
Hoeffding's inequality bounds both tails of \(\sqrt n(C-p_n)\), for
\(x>1/\sqrt n\), by \(2\exp[-2(x-1/\sqrt n)^2/m]\). This gives uniform
integrability of its squared values and therefore
\(n\operatorname{Var}(C)\to v\). It does not determine the \(1/n\) mean
drift. \(\square\)

Appendix B.4 relates this proof to classical quantile linearization
\citep{woodruff1952,ghosh1971,franciscofuller1991}.

Lean verifies both limits for \(F(\hat q)\), including the case of zero
limiting variance. The formal proof derives the moment limits from the
cluster model; it does not assume them. Appendix F.1 gives the verified
scope and reproduction instructions.

\subsection{Interpretation and scope}\label{interpretation-and-scope}

Only the variance of a cluster's indicator sum enters the limit. If
members share the same marginal distribution but have different pairwise
correlations, replace \(\rho_I(p)\) by their mean \(\bar\rho_I(p)\). The
same proof then permits arbitrary within-cluster dependence. A negative
mean correlation gives \(D(p)<1\) and can improve first-order precision.
This is a variance comparison, not a general penalty for every form of
clustering.

For fixed cluster size and level, two cluster laws with the same
\(\delta(p)\) have the same limiting variance. Their finite-sample
distributions and higher-order terms can still differ. The theorem is
pointwise in the cluster law and the operating level. It supplies no
uniform approximation error across copulas or as \(p\) approaches zero
or one.

Assumption (A3) identifies the target of the calculation. A future
observation from a sampled cluster can depend on the reference data and
requires a different analysis. Likewise, a change in the test marginal
replaces \(F(\hat q)\) by another CDF evaluated at the cutoff. Appendix
A distinguishes such target changes from sampling dispersion.

Under independent continuous sampling, the exact coverage law is
\(\mathrm{Beta}(k,n+1-k)\). Under independent clusters whose \(m\)
members are identical, it is
\(\mathrm{Beta}(\lceil k/m\rceil,b+1-\lceil k/m\rceil)\)
\citep{ramos2026}. The first-order variance recovers both cases, but it
does not recover all their rank-rounding effects. Zero indicator
correlation at one level does not establish independence or an exact
Beta law. Appendix B.3 describes the separate Beta approximation used in
numerical comparisons.

\subsection{The mean is not determined by indicator
correlation}\label{the-mean-is-not-determined-by-indicator-correlation}

Theorem 1 describes first-order fluctuations; it does not determine the
mean of \(C\) to order \(1/n\). A continuity-corrected normal
calculation (Appendix B.7) suggests the candidate expansion

\[\mathbb E C-p_n=\frac{A(p_n)}n+o(n^{-1}),\qquad
A(p)=\frac{m-1}{2}\{p(1-p)\rho_I'(p)-(2p-1)\rho_I(p)\}.\]

\textbf{This expansion does not hold in general under (A1)--(A3).} Let
each cluster be a random permutation of \((U,U,V,V)\), with independent
uniform \(U,V\). Use independent draws for different clusters. Then
\(m=4\), \(\rho_I(p)=1/3\), and \(D(p)=2\). All three assumptions hold.
The pooled sample repeats each of \(2b\) independent uniforms twice, so

\[\mathbb E C=\frac{\lceil k/2\rceil}{2b+1}.\]

At \(p_0=0.9\), along \(b\) divisible by ten,
\(b(\mathbb E C-p_n)\to+0.025\), whereas the candidate coefficient gives
\(-0.100\). The cluster count has lattice span two. Smooth \(D\) and
positive variance do not remove this lattice effect.

Theorem 1 holds for this construction, with \(D=2\). A second-order mean
term needs further conditions on the cluster law, including its lattice
structure. Appendix B.7 gives the smooth-integral calculation and the
remainder it leaves unbounded; agreement in particular models, such as
the Gaussian computations of Appendix E.2, does not establish the
expansion in general.

\section{Score correlation can miss the coverage
variance}\label{score-correlation-can-miss-the-coverage-variance}

The design effect for a mean uses the correlation of the measured
values. A sample-quantile cutoff has a different source of first-order
error: the count below the population quantile. Substituting score
correlation for indicator correlation can therefore give the wrong
variance. The error can occur in either direction.

\subsection{A zero-correlation
counterexample}\label{a-zero-correlation-counterexample}

Take independent clusters of size two. Within each cluster, draw \(U\)
uniformly on \([0,1]\). Set the pair to \((U,U)\) with probability \(q\)
and to \((U,1-U)\) otherwise, with the mixture choice independent of
\(U\). Both marginals are uniform and the pair is exchangeable. Its
score correlation is \(\rho=2q-1\). For \(p\ge1/2\),

\[\delta(p)=qp+(1-q)(2p-1),\qquad
\rho_I(p)=\frac{q-(1-p)}p.\]

At \(q=1/2\) and \(p=0.9\), score correlation is zero but
\(\rho_I(p)=4/9\). The coverage design effect is therefore
\(D(p)=13/9\). A score-correlation substitution would return one and
miss a 44\% first-order variance increase.

{\def\LTcaptype{none} 
\begin{longtable}[]{@{}
  >{\raggedright\arraybackslash}p{(\linewidth - 8\tabcolsep) * \real{0.2000}}
  >{\raggedright\arraybackslash}p{(\linewidth - 8\tabcolsep) * \real{0.2000}}
  >{\raggedright\arraybackslash}p{(\linewidth - 8\tabcolsep) * \real{0.2000}}
  >{\raggedright\arraybackslash}p{(\linewidth - 8\tabcolsep) * \real{0.2000}}
  >{\raggedright\arraybackslash}p{(\linewidth - 8\tabcolsep) * \real{0.2000}}@{}}
\toprule\noalign{}
\begin{minipage}[b]{\linewidth}\raggedright
Mixture weight \(q\)
\end{minipage} & \begin{minipage}[b]{\linewidth}\raggedright
Score correlation \(\rho\)
\end{minipage} & \begin{minipage}[b]{\linewidth}\raggedright
Indicator correlation \(\rho_I(0.9)\)
\end{minipage} & \begin{minipage}[b]{\linewidth}\raggedright
Indicator design effect
\end{minipage} & \begin{minipage}[b]{\linewidth}\raggedright
Score-based substitute
\end{minipage} \\
\midrule\noalign{}
\endhead
\bottomrule\noalign{}
\endlastfoot
0.75 & 0.50 & 0.7222 & 1.7222 & 1.50 \\
0.50 & 0.00 & 0.4444 & 1.4444 & 1.00 \\
0.30 & \(-0.40\) & 0.2222 & 1.2222 & 0.60 \\
\end{longtable}
}

These are population quantities at \(p=0.9\), not fitted finite-sample
moments. For \(1-p<q<1/2\), score correlation is negative while
indicator correlation is positive. Thus a score-based correction can
suggest improved precision when the coverage variance is inflated.

\subsection{Numerical comparison under a Gaussian
copula}\label{numerical-comparison-under-a-gaussian-copula}

We also compare the first-order standard deviation with simulation under
Gaussian dependence. The reference sample has \(b=50\) clusters of
\(m=4\), so \(n=200\). At target \(p_0=0.90\), the rank is \(k=181\) and
\(p_n=181/201\). Each of 20,000 repetitions evaluates \(C=F(\hat q)\)
directly. This avoids additional noise from a finite test set.

{\def\LTcaptype{none} 
\begin{longtable}[]{@{}
  >{\raggedright\arraybackslash}p{(\linewidth - 10\tabcolsep) * \real{0.1667}}
  >{\raggedright\arraybackslash}p{(\linewidth - 10\tabcolsep) * \real{0.1667}}
  >{\raggedright\arraybackslash}p{(\linewidth - 10\tabcolsep) * \real{0.1667}}
  >{\raggedright\arraybackslash}p{(\linewidth - 10\tabcolsep) * \real{0.1667}}
  >{\raggedright\arraybackslash}p{(\linewidth - 10\tabcolsep) * \real{0.1667}}
  >{\raggedright\arraybackslash}p{(\linewidth - 10\tabcolsep) * \real{0.1667}}@{}}
\toprule\noalign{}
\begin{minipage}[b]{\linewidth}\raggedright
Score correlation \(\rho\)
\end{minipage} & \begin{minipage}[b]{\linewidth}\raggedright
Indicator correlation at \(p_n\)
\end{minipage} & \begin{minipage}[b]{\linewidth}\raggedright
Effective size
\end{minipage} & \begin{minipage}[b]{\linewidth}\raggedright
Simulated sd
\end{minipage} & \begin{minipage}[b]{\linewidth}\raggedright
First-order sd
\end{minipage} & \begin{minipage}[b]{\linewidth}\raggedright
Score-based sd
\end{minipage} \\
\midrule\noalign{}
\endhead
\bottomrule\noalign{}
\endlastfoot
0.00 & 0.0000 & 200.0 & 0.0212 & 0.0212 & 0.0212 \\
0.20 & 0.0798 & 161.4 & 0.0234 & 0.0236 & 0.0268 \\
0.40 & 0.1847 & 128.7 & 0.0265 & 0.0264 & 0.0314 \\
0.60 & 0.3221 & 101.7 & 0.0299 & 0.0297 & 0.0354 \\
0.80 & 0.5135 & 78.7 & 0.0341 & 0.0337 & 0.0390 \\
0.95 & 0.7545 & 61.3 & 0.0388 & 0.0382 & 0.0415 \\
1.00 & 1.0000 & 50.0 & 0.0414 & 0.0423 & 0.0423 \\
\end{longtable}
}

The prediction uses \(\sqrt{p_n(1-p_n)D(p_n)/n}\). The score-based
column substitutes \(\rho\) for \(\rho_I(p_n)\) in the same expression.
At independence, the exact finite-sample law is
\(\mathrm{Beta}(181,20)\). At perfect duplication, it is
\(\mathrm{Beta}(46,5)\). Their standard deviations are 0.0211 and
0.0412, respectively. The difference from the first-order prediction
includes finite-sample rank effects.

\begin{figure}
\centering
\includegraphics[width=1\linewidth,height=\textheight,keepaspectratio,alt={Gaussian simulation and coverage approximations. Left: simulated coverage standard deviations compared with indicator- and score-based predictions. Right: fitted Beta approximations at three dependence levels. The Beta shapes are approximations, not a conclusion of Theorem 1.}]{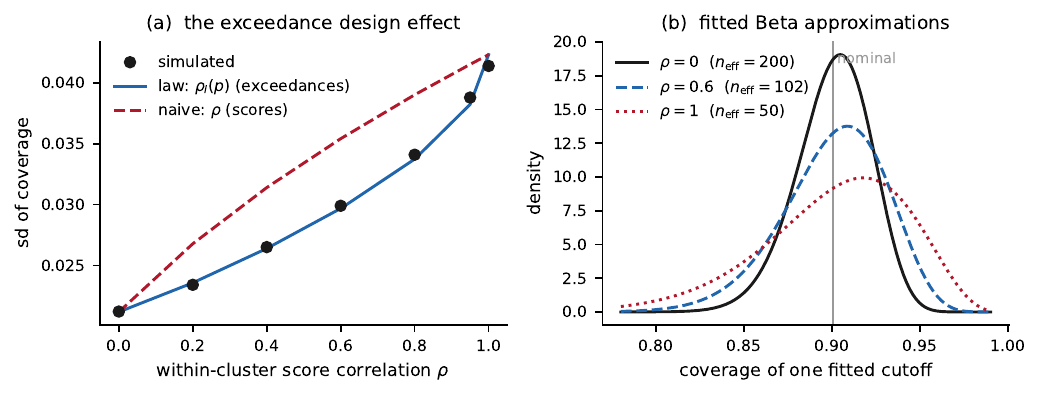}
\caption{Gaussian simulation and coverage approximations. Left:
simulated coverage standard deviations compared with indicator- and
score-based predictions. Right: fitted Beta approximations at three
dependence levels. The Beta shapes are approximations, not a conclusion
of Theorem 1.}\label{fig:law}
\end{figure}

The indicator prediction tracks dispersion in this example. Using score
correlation instead overstates dispersion at the intermediate
correlations tested here. Section 3.1 shows why that direction cannot be
assumed generally. Appendix E reports additional finite-case
comparisons. Neither experiment establishes a uniform approximation
error.

\subsection{A measured score--indicator
contrast}\label{a-measured-scoreindicator-contrast}

SQuAD 2.0 provides an empirical counterpart to Section 3.1
\citep{rajpurkar2018}. We scored its 11,873 development questions with
\texttt{deepset/roberta-base-squad2}. The score is the no-answer logit
minus the best answer-span logit. The questions belong to 1,204
paragraphs within 35 articles. At the paragraph level, the estimated
score ICC is \(-0.0026\). After thresholding at the 90th percentile, the
estimated indicator ICC is \(+0.0640\). The latter exceeds the 95th
percentile of its shuffled-membership null, 0.0074.

The score estimate is near zero; its negative sign does not establish
negative population correlation. The useful contrast is that
thresholding reveals dependence which the score ICC nearly misses.
Paragraphs also share articles, so treating all paragraphs as
independent would require further justification. We report this as a
comparison of correlation estimates, not a validated paragraph-based
effective size. Appendix F identifies the score and measurement
artifacts.

\section{Operating levels and unequal cluster
sizes}\label{operating-levels-and-unequal-cluster-sizes}

\subsection{One dataset can have several effective sample
sizes}\label{one-dataset-can-have-several-effective-sample-sizes}

Changing the cutoff changes the indicator whose correlation controls
coverage variance. For Gaussian scores with correlation \(\rho\), let
\(z_p=\Phi^{-1}(p)\). Then

\[\rho_I(p)=\frac{\Phi_2(z_p,z_p;\rho)-p^2}{p(1-p)},\]

where \(\Phi_2\) is the bivariate standard normal CDF. This is the
classical relation between latent normal and binary correlation
\citep{kraemer1979,donnereliasziw1994}.

For \(n=200\), \(m=4\), and \(\rho=0.6\), the operating level changes
the effective size:

\begin{center}
\begin{tabular}{@{}lllll@{}}
\toprule
Target $p_0$ & Rank baseline $p_n$ & $\rho_I(p_n)$ & $D(p_n)$ & $n_{\mathrm{eff}}(p_n)$ \\
\midrule
0.90 & 0.9005 & 0.3221 & 1.97 & 101.7 \\
0.95 & 0.9502 & 0.2738 & 1.82 & 109.8 \\
0.99 & 0.9900 & 0.1792 & 1.54 & 130.1 \\
\bottomrule
\end{tabular}
\end{center}

In this model, the relative variance inflation decreases toward the
upper tail. That does not make a rare-event probability easy to
estimate: its relative error can remain large. Nor does the table
establish that indicator correlation decreases with the level for every
cluster law. Released data can show a similar decline. In Section 5.3's
document corpus, the estimated indicator correlation falls from 0.135 at
the median to 0.027 at the 95th percentile. At about 1,000 rows drawn as
whole documents, the measured effective size rises from 217 to 621
across those levels.

\begin{figure}
\centering
\includegraphics[width=1\linewidth,height=\textheight,keepaspectratio,alt={Level dependence under a Gaussian copula. Left: indicator correlation at selected levels compared with score correlation. Right: the resulting design effect. The ordering and decline shown here belong to this family; Section 3.1 gives a different ordering.}]{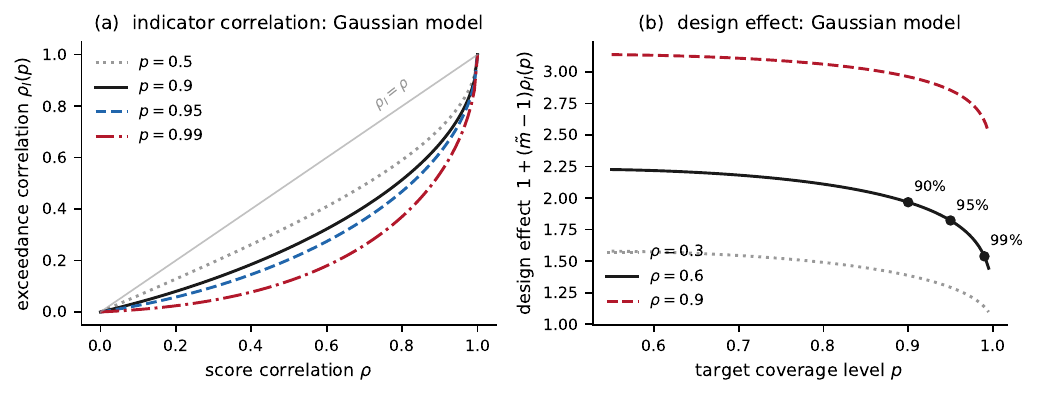}
\caption{Level dependence under a Gaussian copula. Left: indicator
correlation at selected levels compared with score correlation. Right:
the resulting design effect. The ordering and decline shown here belong
to this family; Section 3.1 gives a different
ordering.}\label{fig:level}
\end{figure}

For small \(\rho\), the expansion
\(\rho_I(p)\approx\rho\varphi(z_p)^2/[p(1-p)]\) gives a useful local
calculation \citep{cohen1983}. It can be inaccurate at large correlation
or extreme levels. At \(p=0.99\) and \(\rho=0.8\), the exact indicator
correlation is 6.46 times this approximation. Use the exact Gaussian
expression when that model is justified, or estimate the indicator
correlation directly.

\subsection{Tail limits}\label{tail-limits}

Let \(\mathcal C\) be the pair copula, with diagonal
\(\delta(p)=\mathcal C(p,p)\). The classical upper and lower
tail-dependence coefficients are
\(\lambda_U=\lim_{p\to1}[1-2p+\delta(p)]/(1-p)\) and
\(\lambda_L=\lim_{p\to0}\delta(p)/p\), when these limits exist
\citep{sibuya1960,joe1993}.

\begin{quote}
\textbf{Proposition 1 (tail limits).} Whenever the corresponding
coefficients exist, \[\lim_{p\to1}\rho_I(p)=\lambda_U,\qquad
\lim_{p\to0}\rho_I(p)=\lambda_L.\]
\end{quote}

\textbf{Proof.} The defining identities are
\[\frac{1-2p+\delta(p)}{1-p}=(1-p)+p\rho_I(p),\qquad
\frac{\delta(p)}p=p+(1-p)\rho_I(p).\] Taking the respective limits
proves the claim. \(\square\)

For a Gaussian copula with \(-1<\rho<1\), both tail coefficients vanish.
A copula with \(\lambda_U>0\) instead has \(D(p)\to1+(m-1)\lambda_U\) as
\(p\to1\). This is a limit, not a general lower bound at finite levels.
For example, Section 3.1's mixture of \((U,U)\) and \((U,1-U)\) has
\(\lambda_U=q\). Its indicator correlation \(\rho_I(p)=[q-(1-p)]/p\)
stays below \(q\) at every level \(p<1\).

These identities describe the variance factor's tail behaviour. They do
not extend Theorem 1 to a sequence of target levels approaching one with
sample size. For an application at a fixed level, the quantity to
estimate is \(\rho_I(p)\) at that level.

\subsection{Unequal cluster sizes}\label{unequal-cluster-sizes}

Let cluster \(j\) have size \(m_j\) and let \(n=\sum_jm_j\). Define the
finite-sample size-biased mean

\[\tilde m_b=\frac{\sum_jm_j^2}{\sum_jm_j}
=\bar m_b(1+\mathrm{CV}_b^2),\]

where the cluster-size variance in \(\mathrm{CV}_b\) uses divisor \(b\).
Large clusters receive more weight because they contribute more pairs.

\begin{quote}
\textbf{Proposition 2 (unequal sizes).} Retain (A2)--(A3) and assume:

\textbf{(R1)} Clusters are independent.

\textbf{(R2)} Sizes are fixed by design, or we condition on a
noninformative size profile. Conditional on that profile, all
observations retain marginal CDF \(F\).

\textbf{(R3)} At each level \(p\), every cluster with at least two
members has the same mean pairwise indicator correlation
\(\bar\rho_I(p)\).

\textbf{(R4)} The sizes satisfy \(1\le m_j\le M<\infty\), and
\(\tilde m_b\to\tilde m\) as \(b\to\infty\).

Then Theorem 1 holds with \(D(p_0)=1+(\tilde m-1)\bar\rho_I(p_0)\).
\end{quote}

\textbf{Proof.} On the uniform scale, the pooled count has variance
\[\operatorname{Var}N_b(t)=nt(1-t)[1+(\tilde m_b-1)\bar\rho_I(t)].\]
Bounded cluster counts satisfy the Lindeberg condition when the limiting
normalized variance is positive. For an interval of probability \(h\),
the total increment variance divided by \(n\) is at most \(Mh\). The
count-inversion argument in Section 2.2 therefore applies to this
independent triangular array. If the limiting variance is zero,
Chebyshev's inequality gives the degenerate limit. The bounded-size
concentration argument in Appendix B.9 supplies convergence of moments.
\(\square\)

The common pair-copula diagonal is continuous on interior levels, as in
Appendix B.2. If every cluster is a singleton, the formula reduces
directly to independence. The factor \(\tilde m_b\) is classical
\citep{moulton1986,katz1993}; the extension applies it to coverage
variability. When the common indicator correlation is positive, using
\(\bar m_b\) in its place understates the first-order variance. At
negative correlation, the comparison reverses.

Informative sizes require a separate analysis. They can change both the
variance and the population represented by a pooled row. The simple
substitution need not overestimate or underestimate uncertainty in a
fixed direction. A finite observed maximum size also does not establish
(R4) along an asymptotic sequence. Appendix A explains the
target-population distinction.

Appendix E.3 gives a numerical example: the plain mean understates
standard deviation by 29\% in one uneven-size profile.

\section{Illustrations on released
data}\label{illustrations-on-released-data}

The released process-reward data of Park and colleagues provide an
example of clustered reference observations \citep{park2025}. The
dataset contains 25,028 prefix rows from 500 questions. Prefixes from
the same question share a source, and some also share a reasoning
trajectory. We use the question as the resampling cluster.

The released variable, \texttt{success\_prob}, is a proxy rather than
the nonconformity score used by the original calibration procedure. It
takes nine values, with 67\% of its mass at zero. The original
quantile-head residual is not released. Our measurements therefore
describe this proxy and the stated resampling procedure, not the
deployed system's coverage.

\subsection{Procedure and
measurements}\label{procedure-and-measurements}

We compare whole-question resampling with an independent-row resampling
control. Each repetition samples 500 questions with replacement, pools
their rows, and estimates a cutoff. It evaluates coverage against the
full released pool's empirical CDF, giving each prefix equal weight. The
independent-row control draws a row count from the same resampled size
profiles, so sample size varies in both arms. The main comparison uses
4,000 repetitions.

The randomized arm adds independent uniform jitter of width 0.124 to
each released score once, before resampling. This preserves ordering
between the proxy's atoms, which are 0.125 apart. The empirical
reference CDF and bootstrap both use those augmented scores. Conditional
on that realized pool, the bootstrap distribution remains discrete.
Continuous randomization defines a different underlying procedure; this
empirical bootstrap does not itself satisfy the continuous-population
theorem exactly (Appendix B.1). The operating level is near 0.891, the
proxy's reported level nearest 90\%.

{\def\LTcaptype{none} 
\begin{longtable}[]{@{}
  >{\raggedright\arraybackslash}p{(\linewidth - 2\tabcolsep) * \real{0.5000}}
  >{\raggedright\arraybackslash}p{(\linewidth - 2\tabcolsep) * \real{0.5000}}@{}}
\toprule\noalign{}
\begin{minipage}[b]{\linewidth}\raggedright
Quantity
\end{minipage} & \begin{minipage}[b]{\linewidth}\raggedright
Measurement
\end{minipage} \\
\midrule\noalign{}
\endhead
\bottomrule\noalign{}
\endlastfoot
Rows / question clusters & 25,028 / 500 \\
Cluster size: minimum / median / maximum & 8 / 44 / 135 \\
Mean / size-biased mean cluster size & 50.06 / 61.29 \\
Raw-score ICC & 0.599 \\
Indicator ICC at the reported proxy level & 0.495 \\
Plug-in design effect & 30.8 \\
Plug-in effective size & 812 \\
Ratio of coverage standard deviations, randomized proxy & 4.46× \\
Effective size inferred from randomized-proxy dispersion & About
1,300 \\
Ratio of coverage standard deviations, raw atomic proxy & 1.09× \\
\end{longtable}
}

The randomized-proxy 5th--95th percentile interval is \([0.876, 0.905]\)
under whole-question resampling, compared with \([0.888, 0.894]\) under
independent-row resampling. These are bootstrap distributions of
coverage, not confidence intervals for the original system's deployed
coverage. The 4.46 ratio refers to standard deviations; the displayed
interval endpoints are rounded. A permuted-membership control matches
the independent-row dispersion to 0.2\%. Appendix E.4 reports a
singleton control and a descriptive Beta fit to 8,000 bootstrap
repetitions.

The raw and randomized scores give different dispersion ratios because
ties change the cutoff's sampling behaviour. Neither ratio estimates the
inflation for the unreleased calibration score. In particular, the
randomized result cannot be described as a deployment failure rate for
the original system.

\subsection{Where the simple plug-in
fails}\label{where-the-simple-plug-in-fails}

Cluster size and mean proxy score are associated: their Spearman
correlation is about \(-0.42\). Thus the noninformative-size model in
Proposition 2 is not justified for this dataset. Its plug-in predicts a
standard-deviation ratio of \(\sqrt{30.8}=5.55\), compared with 4.46 in
the bootstrap. The overestimate is a property of this example, not a
general consequence of informative sizes.

A synthetic comparison uses the same size profile with sizes independent
of scores. Its plug-in-to-measured ratio is 0.98. Restricting the
released data to clusters of 40--60 rows reduces that ratio to 1.05.
These checks support size--score coupling as one explanation for the
discrepancy. They do not isolate a unique cause or validate all
assumptions of Proposition 2.

The target population also matters. Changing from equal weight per
prefix to equal weight per question moves coverage from 0.891 to 0.841.
That 4.99-point difference describes a change of target, not sampling
dispersion. Its direction for the original nonconformity score cannot be
inferred from the proxy alone.

This illustration supports a practical distinction: score ICC, indicator
ICC, and observed cutoff dispersion are separate measurements. It also
shows why reporting a plug-in effective size requires the sampling unit,
target population, and uncertainty calculation.

\subsection{A specified score on a document
corpus}\label{a-specified-score-on-a-document-corpus}

The process-reward example lacks the deployed score. A second example
has it. PASC calibrates a named-entity recognizer on 1,000 CoNLL-2003
examples, using the nonconformity score of its Eq. (10) with the
\texttt{dslim/bert-base-NER} model \citep{pasc2026}. CoNLL-2003 ships
its document boundaries. PASC does not state whether its calibration
examples were drawn as sentences or as whole documents.

We computed that score for all 14,041 training sentences and grouped
them by document. The 946 documents have 3 to 197 sentences each, with
mean 14.84 and size-biased mean 26.37. We treat this pool as the
population, with each sentence weighted equally. One fifth of the scores
equal zero. Every cutoff below lies above that atom and ties with a
single row.

Two designs draw about 1,000 rows, repeated 4,000 times. The first draws
sentences independently from the pool. The second draws whole documents
until the next one would exceed 1,000 rows. Each draw fits the
split-conformal cutoff, and the pool's empirical distribution gives its
coverage. The measured effective size is \(p(1-p)\) divided by the
variance of that coverage.

{\def\LTcaptype{none} 
\begin{longtable}[]{@{}
  >{\raggedright\arraybackslash}p{(\linewidth - 10\tabcolsep) * \real{0.1667}}
  >{\raggedright\arraybackslash}p{(\linewidth - 10\tabcolsep) * \real{0.1667}}
  >{\raggedright\arraybackslash}p{(\linewidth - 10\tabcolsep) * \real{0.1667}}
  >{\raggedright\arraybackslash}p{(\linewidth - 10\tabcolsep) * \real{0.1667}}
  >{\raggedright\arraybackslash}p{(\linewidth - 10\tabcolsep) * \real{0.1667}}
  >{\raggedright\arraybackslash}p{(\linewidth - 10\tabcolsep) * \real{0.1667}}@{}}
\toprule\noalign{}
\begin{minipage}[b]{\linewidth}\raggedright
Level \(p\)
\end{minipage} & \begin{minipage}[b]{\linewidth}\raggedright
Indicator correlation
\end{minipage} & \begin{minipage}[b]{\linewidth}\raggedright
Design effect
\end{minipage} & \begin{minipage}[b]{\linewidth}\raggedright
Sentence draws
\end{minipage} & \begin{minipage}[b]{\linewidth}\raggedright
Document draws {[}95\% band{]}
\end{minipage} & \begin{minipage}[b]{\linewidth}\raggedright
Ratio
\end{minipage} \\
\midrule\noalign{}
\endhead
\bottomrule\noalign{}
\endlastfoot
0.50 & 0.135 & 4.42 & 1,002 & 217 {[}208, 227{]} & 4.62× \\
0.80 & 0.083 & 3.09 & 1,014 & 326 {[}312, 341{]} & 3.11× \\
0.90 & 0.037 & 1.95 & 1,019 & 485 {[}465, 507{]} & 2.10× \\
0.95 & 0.027 & 1.69 & 1,000 & 621 {[}595, 649{]} & 1.61× \\
\end{longtable}
}

The correlation and design-effect columns come from the full pool, using
the cluster-count estimator of Section 6.1 in the unequal-size form of
Appendix B.5. Each correlation clears a permutation null that shuffles
document labels, by factors from 6.3 at \(p=0.95\) to 34 at \(p=0.5\).
Sentence draws give an effective size near 1,000, as independent draws
must, which checks the resampling harness. At PASC's operating levels of
80\%, 90\% and 95\%, the sampling unit changes the effective size by
factors of 3.1, 2.1 and 1.6.

Proposition 2 with the count estimate predicts the document-draw
effective sizes as 217, 311, 497 and 575. Each prediction is within 8\%
of the measurement. A one-way ANOVA estimate of the indicator
correlation gives 0.063 at the 90\% level and 0.050 at the 95\% level.
With those values the prediction falls to 372 and 428, which understates
the measured sizes by 23\% and 31\%. The two estimators weight clusters
differently, and the weightings disagree when the correlation varies
between clusters (Appendix B.5). Here the direct measurement agrees with
the count estimator.

The size assumption is doubtful at the lower levels. The Spearman
correlation between document size and the document's rate below the
cutoff is \(+0.15\) at \(p=0.5\) and \(+0.16\) at \(p=0.8\), with
\(p\)-values below \(10^{-5}\). At the 90\% and 95\% levels it is
\(+0.02\) and \(-0.04\). That does not establish noninformative sizes.
This example uses PASC's score and model but not its calibration split,
which is not reported. It does not explain the coverage PASC reports on
its own test set.

Appendix C gives a further indicator example using health-survey
measurements.

\section{Using the result}\label{using-the-result}

\subsection{A reporting procedure}\label{a-reporting-procedure}

\begin{enumerate}
\def\labelenumi{\arabic{enumi}.}
\tightlist
\item
  \textbf{Define the reference unit and the future observation.} State
  what forms a cluster and why clusters can be treated as independent.
  Specify whether the target weights observations or clusters equally.
  If the reference and test populations differ, a dispersion correction
  alone does not repair coverage.
\item
  \textbf{Evaluate dependence at the operating level.} Form the
  threshold indicators and estimate their within-cluster correlation.
  Report the target level, cluster count, size profile, and rare-class
  count. Correlation between numerical scores is not a substitute.
\item
  \textbf{Report uncertainty in the estimate.} Under an appropriate
  cluster model, an indicator ICC or a cluster bootstrap can estimate
  dispersion. Resample independent clusters, rather than individual
  rows, for the latter. Few clusters, uneven sizes, and rare indicators
  can make either estimate unstable
  \citep{cameronmiller2015,fieldwelsh2007}. Random re-splitting of
  pooled calibration and test scores checks a different sampling
  procedure. It does not establish exchangeability under the original
  split; see Section 6.4.
\item
  \textbf{State what the effective size means.} Under Theorem 1 or
  Proposition 2, \(n/D(p)\) describes first-order coverage variance.
  Report it alongside the nominal size and an uncertainty interval. A
  fitted normal or Beta approximation introduces additional
  approximation error.
\item
  \textbf{Repeat the calculation for each operating level.} The same
  reference sample can have different effective sizes at different
  cutoffs. Do not infer tail behaviour from score correlation or from
  one measured level.
\end{enumerate}

For equal-size iid clusters, we can estimate the variance directly from
counts. At the fitted cutoff \(\hat q\), count how many observations in
each cluster fall below it:

\[\hat X_j=\sum_{r=1}^m\mathbb1\{S_{j,r}\le\hat q\},\qquad
\bar X=b^{-1}\sum_{j=1}^b\hat X_j.\]

Compute the sample variance of those counts and divide by \(m\):

\[\hat v=\frac{1}{m(b-1)}\sum_{j=1}^b(\hat X_j-\bar X)^2.\]

Then \(\hat v/n\) estimates the coverage variance. Under Theorem 1,
\(\hat v\) converges in probability to \(p_0(1-p_0)D(p_0)\). Appendix
B.10 proves this even though the same sample supplies the cutoff and the
counts. For \(\hat v>0\), the corresponding estimates are
\(\hat D=\hat v/[p_0(1-p_0)]\) and \(\hat n_{\mathrm{eff}}=n/\hat D\).
This calculation avoids estimating each pairwise correlation separately.
It does not require a positive score density.

With few clusters, assess uncertainty by resampling whole clusters and
recomputing the cutoff and counts each time. Consistency does not
establish finite-sample accuracy or bootstrap validity for every cluster
law. For unequal or informative sizes, the equal-size estimator above
needs a separate justification. With cluster sizes \(m_j\) and
\(\hat p=n^{-1}\sum_j\hat X_j\), the covariance sum of Appendix B.5
suggests

\[\hat v=\frac{b}{b-1}\cdot\frac1n\sum_{j=1}^b(\hat X_j-m_j\hat p)^2,\]

which reduces to the estimator above when all sizes equal \(m\).
Appendix B.10 does not cover this form. Section 5.3 compares it with
direct resampling on one corpus; that comparison is evidence, not a
proof. A null association between size and mean score does not establish
noninformative sizes; check indicator rates at the intended levels as
well.

\subsection{Dispersion and guarantees answer different
questions}\label{dispersion-and-guarantees-answer-different-questions}

A variance approximation describes variation across fitted cutoffs. It
does not ensure that any particular cutoff attains the target coverage.
Nor does substituting an estimated \(n_{\mathrm{eff}}\) into an
independent-sample formula automatically give a valid guarantee.

A separate finite-sample bound follows directly from independent cluster
counts. Suppose sizes \(m_j\) are fixed and all observations have the
same continuous marginal \(F\). For a target \(p\) and failure
probability \(\eta\in(0,1)\), define

\[\varepsilon=\sqrt{\frac{\log(1/\eta)\sum_jm_j^2}{2n^2}},\qquad
k_\eta=\lceil(n+1)(p+\varepsilon)\rceil.\]

Use the cutoff at rank \(k_\eta\), or an infinite cutoff if
\(k_\eta>n\). Then \(\mathbb P\{F(S_{(k_\eta)})\ge p\}\ge1-\eta\), with
the infinite-cutoff convention understood. Appendix B.9 proves this by
Hoeffding's inequality on cluster counts. No indicator-correlation
estimate is required.

The count governing this worst-case bound is
\(n^2/\sum_jm_j^2=n/\tilde m_b\). It differs from the variance-based
effective size. The bound permits arbitrary dependence within clusters,
which can make it conservative. Section 2's asymptotic law and this
finite-sample guarantee should therefore be reported separately.

\subsection{Collecting more
observations}\label{collecting-more-observations}

Under a fixed positive indicator correlation, the equal-size variance
expression gives \(n_{\mathrm{eff}}=bm/[1+(m-1)\rho_I(p)]\). The algebra
suggests diminishing returns from enlarging existing clusters. It does
not extend the fixed-\(m\) theorem to a limit with \(m\to\infty\) and
\(b\) fixed. Dependence can also change with the collection procedure or
cluster size. Compare additional clusters with additional observations
within clusters using measured costs and dependence over the proposed
range.

For a finite test set, its observed coverage proportion has additional
sampling noise. Dependent test observations require their own variance
analysis. Theorem 1 concerns the population coverage of the fitted
cutoff, before that test-set noise is added.

\subsection{What a random re-split
checks}\label{what-a-random-re-split-checks}

PASC gives a concrete example \citep{pasc2026}, Appendix A.1, Table 7.
It pools 1,000 calibration scores and 500 test scores, then repeats a
random split 200 times. For its named-entity stage at target 0.950, the
mean re-split coverage is 0.950. The original test coverage is 0.906. At
targets 0.900 and 0.800, the corresponding pairs are 0.902 versus 0.860
and 0.798 versus 0.786.

Conditional on the pooled scores, random assignment supplies
exchangeability for the new split. Agreement with the target therefore
checks rank calibration under that randomized procedure. It does not
verify the sampling assumptions behind the original split. The
discrepancy alone neither identifies clustering as its cause nor refutes
a marginal coverage theorem. Section 5.3 separately shows how the
sampling unit changes dispersion for the same score and model. These
table references concern PASC's original arXiv version.

\section{Relation to existing
methods}\label{relation-to-existing-methods}

\subsection{Clustered quantiles and indicator design
effects}\label{clustered-quantiles-and-indicator-design-effects}

Woodruff's quantile intervals invert uncertainty on the CDF scale
\citep{woodruff1952}. Kish explains this route in Section 12.9,
``Standard errors for medians and quantiles'' \citep{kish1965} (p.~495).
He estimates uncertainty in the proportion below the sample quantile,
then inverts the cumulative distribution. For complex samples, he
explicitly requires a variance formula appropriate to the sampling
design. Thus the indicator approach to clustered quantiles predates
modern conformal prediction. Francisco and Fuller establish quantile
asymptotics under complex survey designs \citep{franciscofuller1991}.
Under positive density, their familiar linearization connects quantile
error with threshold-indicator error. Applying the population CDF
cancels the density factor (Appendix B.4). Theorem 1 gives a direct
proof on the coverage scale for independent clusters. It does not
require the additional design conditions of a general survey theorem.

The design effect itself is classical \citep{kish1965}. Katz and
colleagues combine a binary indicator correlation with the size-biased
mean cluster size \citep{katz1993}. The unequal-size factor also appears
in Moulton's regression analysis, which credits Campbell (1977)
\citep{moulton1986}. We use that variance identity for a sample-quantile
cutoff's coverage. We claim no new design-effect formula for a fixed
binary outcome.

Online experimentation supplies a close applied precedent. Deng and
colleagues estimate clustered quantile variance using indicators at the
quantile \citep{deng2018}. Yao and colleagues use this construction in
their quantile-treatment-effect algorithm \citep{yao2024}. These methods
already establish the importance of indicator variation rather than
raw-score variation. Our emphasis is the resulting distribution of
population coverage across reference samples, its explicit
indicator-correlation parameter, and its dependence on the operating
level.

\subsection{Conditional coverage under
dependence}\label{conditional-coverage-under-dependence}

The independent-sample Beta law and training-conditional guarantees are
established in conformal prediction \citep{vovk2012,bianbarber2022}.
Wieczorek connects conformal prediction with survey sampling and asks
for effective-size estimates under non-exchangeable designs
\citep{wieczorek2023}. Hierarchical conformal methods change the
calibration procedure to obtain validity for grouped data
\citep{dunn2023,leebarberwillett2023}. Our question concerns the
variability of coverage when the cutoff remains a pooled sample
quantile.

Ramos and colleagues are the closest coverage-law predecessor
\citep{ramos2026}. They express coverage through transformed order
statistics, give a counting representation, and derive the exact law for
perfectly duplicated clusters. Their dependent-sequence analysis also
uses a sum of indicator covariances. Where the corresponding stationary
representation and regularity assumptions apply, that covariance sum
gives the same first-order variance as our cluster calculation.

Their quantitative mixing result invokes conditions (C.1)--(C.5) of
Lahiri and Sun \citep{lahirisun2009}. At perfect duplication,
conditioning on a cluster-mate reveals whether an observation is below
the quantile. That violates the nondegeneracy requirement in (C.5). Our
independent-cluster proof includes that endpoint without a density or a
quantitative approximation rate. This is a difference in assumptions and
conclusions. It is not a claim of uniform approximation accuracy over
all dependence structures.

Lou and Luo propose a further design-effect adjustment to a weighting
effective size \citep{louluo2026}. Their spatial-neighbour correlation
differs from the threshold-indicator correlation considered here.
Section 3 therefore compares indicator and score corrections as
mathematical alternatives; it does not attribute the raw-score
substitution to their procedure.

\subsection{Operating levels and
copulas}\label{operating-levels-and-copulas}

Dependence of binary ICC on prevalence is established in cluster-trial
design \citep{crespi2011}. The relation between continuous and
dichotomized correlations is also classical
\citep{kraemer1979,donnereliasziw1994,cohen1983}. The curve in Section 4
is related algebraically to the copula tail-concentration function
\citep{venter2002,durante2015} and quantile-dependence measurements
\citep{patton2013}. Its endpoint limits follow from the definitions of
tail dependence \citep{sibuya1960,joe1993}.

These results supply the ingredients of our interpretation. Reading the
indicator correlation at the chosen cutoff connects them to coverage
precision. The contribution is the explicit connection and its
consequences, rather than priority for the individual ingredients.

\section{Scope and further questions}\label{scope-and-further-questions}

The stated mathematical results are proved and mapped to Lean
declarations in Appendix F. A further practical question is accuracy
with few clusters. Theorem 1 gives a large-sample variance formula, and
Section 6 gives a consistent estimator. Neither specifies how many
clusters make a normal approximation accurate. Section 6.2 provides a
finite-sample guarantee when such a guarantee is needed, though it can
be conservative. A sharper error bound would help practitioners choose
between these approaches.

The theorem assumes independent clusters and a fixed target fraction.
Dependent clusters and target fractions that approach zero or one
require further analysis. The unequal-size result also requires the size
assumptions in Proposition 2. These assumptions should be checked for
each application.

The separate question of mean bias remains open beyond special cases.
Section 2.4 shows that the indicator correlation alone does not
determine it. Appendix F records the scope of the Lean verification and
the numerical checks.

\clearpage
\appendix
\makeatletter
\renewcommand{\@seccntformat}[1]{\ifnum\pdfstrcmp{#1}{section}=0 Appendix~\csname the#1\endcsname:\quad\else\csname the#1\endcsname\quad\fi}
\makeatother

\section{Informative sizes and target
populations}\label{informative-sizes-and-target-populations}

\subsection{A row target differs from a cluster
target}\label{a-row-target-differs-from-a-cluster-target}

Suppose clusters have random sizes \(M_j\), and let \(G_j(t)\) be the
fraction of cluster \(j\) below \(t\). Under iid clusters with
\(0<\mathbb E M_j<\infty\), the pooled empirical CDF converges to

\[F_{\mathrm{row}}(t)=\frac{\mathbb E[M_jG_j(t)]}{\mathbb E M_j}.\]

Sampling a cluster uniformly and then a member uniformly instead gives
\(F_{\mathrm{cluster}}(t)=\mathbb E G_j(t)\). Their difference is
\(\operatorname{Cov}(M_j,G_j(t))/\mathbb E M_j\). Thus informative sizes
can change which distribution a pooled quantile estimates. If the future
observation follows the same row-weighted law, that particular target
mismatch is absent. This does not restore the common-correlation
variance formula automatically.

Section 5's reweighting comparison describes this distinction in a
released proxy dataset. It does not establish a bias for an unspecified
deployment target. Known sampling weights can address some target
differences; their estimation and variance require their own analysis.

\subsection{Unknown selection can hide a target
difference}\label{unknown-selection-can-hide-a-target-difference}

If an observation with score \(s\) is retained with probability
\(\pi(s)\), the observed CDF is

\[F_\pi(t)=\frac{\int_{-\infty}^t\pi(s)\,dF(s)}{\mathbb E\pi(S)}.\]

A quantile of this distribution need not have coverage \(p\) under the
unselected population \(F\) \citep{pfeffermann1998,bonnery2012}. A
variance correction does not identify the target law from selected
observations alone.

For example, consider a fixed-size sample of accepted scores with no
rejection information. One model has uniform population scores on
\([0,1]\) and no selection. A second has density \(f_a(s)=1+a(2s-1)\),
for \(0<a<1\), and retention probability \((1-a)/f_a(s)\). Both generate
uniform accepted scores. Yet at cutoff \(p\), the second population has
coverage \(p-ap(1-p)\). At \(p=0.9\) and \(a=0.9\), the difference is
8.1 percentage points. It is an example of non-identification under
unknown selection, not a universal size or sign of bias.

Known selection probabilities with adequate positivity permit
inverse-probability weighting \citep{vardi1985,gillvardiwellner1988}.
Bounds on likelihood ratios permit sensitivity analyses
\citep{jinrencandes2022}. An independent, unfiltered reference sample
provides another source of information about the target. These are
separate estimation problems; none follows from substituting an
effective size into Theorem 1.

\subsection{Label provenance can change the reference
population}\label{label-provenance-can-change-the-reference-population}

A gate calibrated on labelled errors uses the score distribution among
those labelled errors. Replacing ground truth with model-generated
labels can change which observations enter that distribution. The
overall label-error rate does not determine the resulting cutoff: the
score--label relationship also matters. Random label corruption and
score-dependent mistakes therefore need not move a cutoff in the same
direction. No general sign rule follows without assumptions on that
relationship \citep{einbinder2022,sesia2025}.

The practical distinction is between identifying the target and
estimating dispersion around it. Known selection probabilities can
support weighting, provided every relevant part of the target remains
observable. Such weights change the variance calculation and can
increase it substantially. A bound on selection strength instead
supports a sensitivity interval. Fresh observations from the target
avoid that selection model, but require new data. The choice depends on
what information is available; none is a consequence of an
effective-size substitution.

\section{Technical details}\label{technical-details}

\subsection{Atomic scores and
randomization}\label{atomic-scores-and-randomization}

Continuity in (A2) concerns each marginal, not the absence of ties
within a cluster. Perfect within-cluster duplication of a continuous
random variable is allowed. A marginal with atoms is outside Theorem 1.

Independent continuous jitter can produce a continuous augmented score,
such as \(S+\varepsilon U\). Calibration and testing must both use that
same randomized rule. Its coverage is not the coverage of the original
atomic procedure. For Section 5's proxy, the measured dispersion ratios
are 1.09 for the raw score and 4.46 after randomization. These
measurements cannot determine the ratio for an unreleased residual
score.

A duplication mechanism need not create marginal atoms. For a pair that
is identical with probability \(q\) and independent otherwise, with
continuous common marginals, \(\delta(p)=qp+(1-q)p^2\) and
\(\rho_I(p)=q\) at every level.

\subsection{Fixed and finite-sample
levels}\label{fixed-and-finite-sample-levels}

The CLT is stated at the fixed target \(p_0\). Centering at
\(p_n=k/(n+1)\) gives the same limit because
\(\sqrt n|p_n-p_0|\le\sqrt n/(n+1)\to0\).

The copula diagonal is 2-Lipschitz, so \(\rho_I\) is continuous on
\((0,1)\). Consequently, \(\rho_I(p_n)\to\rho_I(p_0)\) without an
additional smoothness assumption. A finite average of pair-copula
diagonals has the same continuity property. Numerical first-order
comparisons evaluate their variance factors at \(p_n\).

\subsection{The Beta approximation}\label{the-beta-approximation}

For \(n_{\mathrm{eff}}>1\), consider

\[C\approx\mathrm{Beta}\!\left(p_n(n_{\mathrm{eff}}-1),(1-p_n)(n_{\mathrm{eff}}-1)\right).\]

This distribution has mean \(p_n\) and variance
\(p_n(1-p_n)/n_{\mathrm{eff}}\). It matches the baseline mean and
first-order variance; it need not match the actual finite-sample
moments. A Gaussian limit does not select a Beta distribution or justify
its tail probabilities.

Even under independent continuous sampling, this approximation differs
from the exact law \(\mathrm{Beta}(k,n+1-k)\). That exact law has
variance \(p_n(1-p_n)/(n+2)\). The difference is a finite-sample term,
not a failure of the first-order theorem. Zero indicator correlation at
a single level does not imply that the exact Beta law applies. For
example, use Section 3.1's pair mixture with \(q=1/3\) and target
\(p_0=2/3\). The indicator correlation is zero, and the ceiling rule
selects the maximum of the two scores. Its CDF at \(1/2\) is \(1/6\),
whereas the independent-sample \(\mathrm{Beta}(2,1)\) CDF is \(1/4\).
Lean checks this finite-sample counterexample.

Under dependence, both the finite-sample shape and mean can differ from
the approximation. Any use of its percentiles needs separate validation
for the intended model and sample size.

\subsection{Relation to quantile
linearization}\label{relation-to-quantile-linearization}

For independent clusters of fixed size with common CDF \(F\), suppose
\(F(q_p)=p\in(0,1)\). Assume \(F\) is differentiable at \(q_p\) with
derivative \(f(q_p)>0\). With the prescribed ceiling rank, the Bahadur
representation is

\[\hat q-q_p=-\frac{\hat F(q_p)-p}{f(q_p)}+o_P(n^{-1/2}).\]

Lean checks this representation for the actual pooled quantile and
empirical CDF. Applying \(F\) cancels the density factor
\citep{ghosh1971,franciscofuller1991}. This explains the connection to
classical quantile inference. The count-inversion proof in Section 2.2
establishes the coverage result directly, without borrowing unverified
design conditions or requiring a positive density.

\subsection{The unequal-size variance
identity}\label{the-unequal-size-variance-identity}

Let \(X_j(t)\) be cluster \(j\)'s count on the uniform scale. If every
cluster has the same mean pairwise indicator correlation
\(\bar\rho_I(t)\), then

\[\begin{aligned}
\operatorname{Var}N_b(t)
&=t(1-t)\sum_j\{m_j+m_j(m_j-1)\bar\rho_I(t)\}\\
&=nt(1-t)[1+(\tilde m_b-1)\bar\rho_I(t)].
\end{aligned}\]

This is the classical size-biased substitution
\citep{moulton1986,katz1993}. Proposition 2 states the conditions needed
to transfer it from counts to cutoff coverage. When correlations differ
between clusters, the weighted covariance sum remains available, but a
single common ICC requires an additional assumption.

\subsection{Unequal sizes and the
mean}\label{unequal-sizes-and-the-mean}

The variance identity in Appendix B.5 does not establish a second-order
mean expansion. Replacing \(m\) with \(\tilde m_b\) in the candidate
coefficient of Section 2.4 does not avoid its failure at equal sizes. A
valid expansion needs a declared cluster model and explicit lattice
conditions. The numerical comparisons in Appendix E.2 concern specified
Gaussian models only.

\subsection{The smooth-integral
calculation}\label{the-smooth-integral-calculation}

For a specified independent-cluster model,
\(\mathbb E C=\int_0^1\mathbb P\{N_b(t)\le k-1\}\,dt\) exactly. Write
\(V(t)=t(1-t)D(t)\). A continuity-corrected normal approximation to that
count probability gives, formally,

\[\mathbb E C\approx\frac{k-1/2}{n}+\frac{V'(p_n)}{2n}.\]

Since \((k-1/2)/n-p_n=(p_n-1/2)/n\), its coefficient relative to \(p_n\)
is

\[A(p)=\frac12\frac{d}{dp}\{p(1-p)[D(p)-1]\}.\]

For fixed \(m\), this expands to the candidate coefficient in Section
2.4. The leading formal Edgeworth skewness term integrates to zero
because \(\int_{-\infty}^{\infty}\varphi(z)(z^2-1)\,dz=0\). These
manipulations do not bound the integrated difference between the exact
count CDF and its approximation. Lattice corrections can contribute at
order \(1/n\), as Appendix B.8 demonstrates. We do not claim that
smoothness, positive variance, or span-one aperiodicity alone has been
proved sufficient here.

Ahanaf Hasan Ariq carried this calculation further in unpublished work.
He treated the smooth integral as a deterministic object and derived its
limit under stated regularity conditions. He also measured its gap to
the exact expectation in a Gaussian model. That gap is the remainder
left unbounded above, and bounding it remains open.

Even where a valid expansion exists, its sign needs the full
coefficient. For \(p>1/2\), positive \(\rho_I\) and nonpositive
\(\rho_I'\) make it negative. A positive derivative alone does not make
it positive; the required inequality is
\(p(1-p)\rho_I'(p)>(2p-1)\rho_I(p)\).

\subsection{Details of the duplicated-pair
counterexample}\label{details-of-the-duplicated-pair-counterexample}

Randomly permute \((U,U,V,V)\) within each cluster, with \(U,V\)
independent uniforms. A pair of positions contains the same underlying
uniform with probability \(1/3\). Otherwise the pair is independent.
Thus \(\delta(p)=p/3+2p^2/3\) and \(\rho_I(p)=1/3\).

The pooled rank \(k\) corresponds to rank \(\lceil k/2\rceil\) among
\(2b\) independent uniforms. Its expectation is therefore
\(\lceil k/2\rceil/(2b+1)\). At \(p_0=0.9\) and \(b\) divisible by ten,
\(k=3.6b+1\) is odd, giving

\[\mathbb E C=0.9+\frac{0.1}{2b+1},\qquad
p_n=0.9+\frac{0.1}{4b+1}.\]

Multiplying the difference by \(b\) gives the limit \(0.025\). The
candidate coefficient has \(A(0.9)=-0.4\) and \(b/n=1/4\), giving
\(-0.100\) instead. The first-order limit remains valid with \(D=2\).

\subsection{Finite-sample cluster bound and convergence of
moments}\label{finite-sample-cluster-bound-and-convergence-of-moments}

Let the positive cluster sizes be fixed, let \(n=\sum_jm_j\ge1\), and
write \(S_2=\sum_jm_j^2\). Assume independent clusters and continuous
common marginal \(F\). Within a cluster, dependence is unrestricted. On
the uniform scale, set \(N(t)=\sum_{j,r}\mathbb1\{U_{j,r}\le t\}\). Then
\(\mathbb E N(t)=nt\), and each cluster contributes a count in
\([0,m_j]\). Hoeffding therefore gives, for \(s>0\),
\[\mathbb P\{N(t)-nt\ge s\}\le \exp(-2s^2/S_2),\] with the same bound
for the lower tail.

For an integer rank \(1\le k\le n\), let \(C=U_{(k)}\) and
\(p_n=k/(n+1)\). The exact count identities hold even with ties:
\[\{C\le t\}=\{N(t)\ge k\},\qquad
  \{C>t\}=\{N(t)\le k-1\}.\] The lower-tail event \(\{C<t\}\) is
contained in the first event. At \(t=p_n-x\), the upper-count deviation
is \(nx+p_n\). At \(t=p_n+x\), the lower-count deviation is
\(nx+1-p_n\). If a threshold leaves \([0,1]\), the corresponding event
has probability zero. Thus, for every \(x>0\),
\[\mathbb P\{|C-p_n|>x\}\le2\exp(-2n^2x^2/S_2).\] For bounded cluster
sizes \(m_j\le M\), \(S_2\le Mn\). Writing \(Z_n=\sqrt n(C-p_n)\) gives
\[\mathbb P\{|Z_n|>y\}\le2e^{-2y^2/M},\qquad
  \mathbb E Z_n^4\le 8\int_0^\infty y^3e^{-2y^2/M}\,dy=M^2.\] The
uniform fourth-moment bound makes \(Z_n^2\) uniformly integrable. There
is also a direct route. Set \(Y_n=Z_n^2\). For \(K\ge0\), integration of
its exponential tail gives \[\mathbb E(Y_n-K)_+\le M e^{-2K/M}.\] Since
\(Y_n\mathbb1\{Y_n\ge K\}\le2(Y_n-K/2)_+\),
\[\sup_n\mathbb E[Y_n\mathbb1\{Y_n\ge K\}]\le2M e^{-K/M}\longrightarrow0.\]
This proves uniform integrability directly; Lean checks both the
equal-size and bounded unequal-size cases. Combined with the
count-inversion CLT, uniform integrability yields convergence of the
first two moments. The limiting Gaussian has mean zero, so
\(n\operatorname{Var}(C)\) converges to its variance. Lean proves both
moment limits and the resulting variance limit for equal-size and
bounded unequal-size clusters. Its proof applies dominated convergence
directly to tail integrals. Lean also checks the exact fourth-moment
bound above. A checked bounded independent-array CLT completes both
limits in Proposition 2. Neither route supplies a second-order
mean-drift expansion.

For a target \(0<p<1\) and failure probability \(0<\eta<1\), choose
\(\varepsilon=\sqrt{S_2\log(1/\eta)/(2n^2)}\) and
\(k=\lceil(n+1)(p+\varepsilon)\rceil\). If \(k\le n\), \(\{C<p\}\)
requires an upper-count deviation of at least \(n\varepsilon\). Its
probability is at most \(\eta\). If \(k>n\), use the infinite threshold,
whose coverage is one. This establishes the one-sided correction in
Section 6.2 without replacing \(n\) by a variance-based effective size.

\subsection{Estimating the coverage variance from cluster
counts}\label{estimating-the-coverage-variance-from-cluster-counts}

Under (A1)--(A3), with fixed \(m\) and \(b\to\infty\), the estimator in
Section 6.1 satisfies

\[\hat v\xrightarrow{P}v=p_0(1-p_0)D(p_0).\]

\textbf{Proof.} Work on the uniform scale and write
\(X_j(t)=\sum_r\mathbb1\{U_{j,r}\le t\}\). The continuous marginal lets
us identify the fitted-cutoff counts with \(X_j(C)\) almost surely.
Theorem 1 gives \(C\to p_0\) in probability. For fixed
\(0<h<\min(p_0,1-p_0)\), on \(|C-p_0|\le h\),

\[\frac1b\sum_j|X_j(C)-X_j(p_0)|
\le\frac1b\sum_j\{X_j(p_0+h)-X_j(p_0-h)\}.\]

The law of large numbers makes the right side converge to \(2mh\). First
let \(b\to\infty\), then \(h\downarrow0\). The average absolute
difference therefore converges to zero in probability. Because
\(0\le X_j(t)\le m\), \(|X_j(C)^2-X_j(p_0)^2|\le2m|X_j(C)-X_j(p_0)|\).
The empirical first and second moments at \(C\) thus have the same
limits as those at \(p_0\). The law of large numbers at that fixed level
gives

\[\hat v\xrightarrow{P}\frac{\operatorname{Var}X_1(p_0)}m=v.\]

By Theorem 1's variance conclusion,
\(\hat v-n\operatorname{Var}(C)\to0\) in probability. If \(v>0\), this
also gives relative consistency:
\((\hat v/n)/\operatorname{Var}(C)\to1\) in probability. At \(v=0\),
only the absolute statement follows. \(\square\)

Lean checks this estimator argument, including the sample-dependent
cutoff and the original-score counts. It also checks the sample-variance
divisor and the absolute and relative consistency statements.

\section{Indicator correlations in
NHANES}\label{indicator-correlations-in-nhanes}

\subsection{An unweighted descriptive
comparison}\label{an-unweighted-descriptive-comparison}

We analyze NHANES 2013--2018 biochemistry measurements for 14,705 adults
aged 18--70. The analysis groups observations into 90 public variance
groups using masked stratum and pseudo-PSU identifiers. These
identifiers do not reveal physical sites, teams, or collection times.
The unweighted estimates describe the released sample, not the national
population under the survey design.

For ten analytes, the procedure compares a one-way ANOVA ICC of log
concentration with the ICC of its threshold indicator. It evaluates four
levels and 200 permutations of cluster labels. A cell passes the
descriptive screen if its indicator ICC exceeds the permutation null's
95th percentile. Ten analytes pass at 0.90, eight at 0.95, seven at
0.975, and five at 0.99. The table keeps the same five analytes at every
level.

\Needspace{9\baselineskip}

{\def\LTcaptype{none} 
\begin{longtable}[]{@{}
  >{\raggedright\arraybackslash}p{(\linewidth - 10\tabcolsep) * \real{0.1667}}
  >{\raggedright\arraybackslash}p{(\linewidth - 10\tabcolsep) * \real{0.1667}}
  >{\raggedright\arraybackslash}p{(\linewidth - 10\tabcolsep) * \real{0.1667}}
  >{\raggedright\arraybackslash}p{(\linewidth - 10\tabcolsep) * \real{0.1667}}
  >{\raggedright\arraybackslash}p{(\linewidth - 10\tabcolsep) * \real{0.1667}}
  >{\raggedright\arraybackslash}p{(\linewidth - 10\tabcolsep) * \real{0.1667}}@{}}
\toprule\noalign{}
\begin{minipage}[b]{\linewidth}\raggedright
Analyte
\end{minipage} & \begin{minipage}[b]{\linewidth}\raggedright
Score ICC
\end{minipage} & \begin{minipage}[b]{\linewidth}\raggedright
\(\hat\rho_I(0.90)\)
\end{minipage} & \begin{minipage}[b]{\linewidth}\raggedright
\(\hat\rho_I(0.95)\)
\end{minipage} & \begin{minipage}[b]{\linewidth}\raggedright
\(\hat\rho_I(0.975)\)
\end{minipage} & \begin{minipage}[b]{\linewidth}\raggedright
\(\hat\rho_I(0.99)\)
\end{minipage} \\
\midrule\noalign{}
\endhead
\bottomrule\noalign{}
\endlastfoot
Sodium & 0.236 & 0.1133 & 0.0732 & 0.0732 & 0.0364 \\
Potassium & 0.104 & 0.0449 & 0.0215 & 0.0149 & 0.0074 \\
Albumin & 0.119 & 0.0307 & 0.0241 & 0.0128 & 0.0049 \\
Creatinine & 0.036 & 0.0136 & 0.0050 & 0.0036 & 0.0032 \\
Bilirubin & 0.114 & 0.0133 & 0.0042 & 0.0046 & 0.0023 \\
\end{longtable}
}

The estimates are lower at 0.99 than at 0.90 for all five analytes. They
do not all decline monotonically: bilirubin rises between 0.95 and
0.975. These are selected descriptive estimates, without simultaneous
uncertainty bands. They illustrate level dependence but do not establish
a general attenuation rule.

Using score ICC in a Gaussian-copula formula also gives analyte-specific
errors. Across the reported cells, predicted-to-measured
indicator-correlation ratios range from 0.39 to 7.03. The Gaussian
prediction is therefore not uniformly conservative on these data. The
analysis screens concentrations below \(10^{-6}\) before taking
logarithms; small coded values otherwise strongly affect the bilirubin
score ICC. That screen and the selection of analytes must accompany the
result.

\subsection{A separate survey-design
calculation}\label{a-separate-survey-design-calculation}

A second calculation estimates weighted CDF variances at observed
thresholds using Taylor linearization. It includes weights, strata,
masked variance units, and the full design for domain estimation. An
independent implementation using samplics 0.4.33 agrees across all 40
analyte--level cells. The largest relative variance difference is below
\(2\times10^{-15}\).

This checks implementation agreement for the stated fixed-threshold
estimator. The thresholds were selected from the data; their selection
uncertainty is not included in this comparison. The variance ratios
combine weighting, stratification, and clustering. They neither isolate
an indicator-ICC design effect nor validate a weighted-quantile coverage
theorem. Recorded measurement ties also prevent treating these
observations as exactly continuous.

\section{Applications and the choice of
estimator}\label{applications-and-the-choice-of-estimator}

\subsection{Where the calculation
applies}\label{where-the-calculation-applies}

Theorem 1 applies to a cutoff that is a sample order statistic under its
sampling assumptions. Examples include empirical reference limits and
latency cutoffs from independent user clusters. A fixed business cutoff
has no quantile-estimation error of this kind. A fitted or smoothed
centile has a different estimator and generally a different variance.

\textbf{Latency cutoffs.} Suppose a team estimates a 95th-percentile
latency cutoff from requests grouped by user. Independent users can
supply the clusters, with the remaining size and marginal assumptions
stated explicitly. The relevant indicators mark requests below the
population 95th percentile. A 99th-percentile cutoff requires a new
calculation at that level. Existing methods for quantile metrics already
use this indicator approach \citep{deng2018,yao2024}.

\textbf{Medical imaging.} Repeated scans or slices from one patient can
share a source of variation. For an empirical score cutoff, independent
patients can supply the clusters. The target must specify whether it
weights patients or images equally (Appendix A). Medical-image conformal
methods face this exchangeability question \citep{lambert2024}. A fitted
volume model or smoothed centile requires the variance of its own
estimator.

\textbf{Value-at-risk.} Christoffersen's forecast checks use exceedance
indicators rather than return correlation \citep{christoffersen1998}.
This is a useful parallel, but serial dependence differs from
independent clusters. Theorem 1 does not justify treating arbitrary
consecutive time blocks as independent. An empirical loss quantile fits
the present framework only when its sampling design meets the stated
cluster assumptions.

\subsection{Fitted centiles use a different influence
function}\label{fitted-centiles-use-a-different-influence-function}

For a regular estimator of \(q_p\), suppose the influence function
\(\psi\) has mean zero and finite second moment. Assume independent iid
clusters of fixed size and a valid asymptotic linear representation. If
\(F\) has a positive continuous density at \(q_p\), the coverage-scale
normal approximation has variance

\[\frac{f(q_p)^2}{bm^2}\operatorname{Var}\!\left(\sum_{r=1}^m\psi(S_{1,r})\right).\]

Convergence of actual rescaled variances additionally requires moment
control. An \(o_P\) remainder alone does not provide it. For an
empirical quantile, \(\psi(s)=[p-\mathbb1\{s\le q_p\}]/f(q_p)\), which
recovers the indicator expression.

For a fitted normal centile on the standard normal scale,
\(\psi_z(x)=x+(z/2)(x^2-1)\). For a jointly standard normal pair with
correlation \(r\), its influence-function correlation is

\[\frac{r+(z^2/2)r^2}{1+z^2/2}.\]

It is generally different from the indicator correlation at
\(p=\Phi(z)\). The estimator, as well as the sampling design and
operating level, determines the relevant dependence quantity.

\section{Additional numerical
comparisons}\label{additional-numerical-comparisons}

\subsection{Matched dependence and nested
clusters}\label{matched-dependence-and-nested-clusters}

Gaussian, \(t_3\), and Clayton copulas tuned to \(\delta(0.9)=0.855\)
have \(\rho_I(0.9)=0.5\). At \(m=4\) and \(b=50\), the first-order
standard deviation is 0.0335. Simulated standard deviations are 0.0331,
0.0338, and 0.0334. This is a finite-case comparison of the common
limiting variance, not evidence of identical finite-sample laws.

Nested Gaussian clusters also test the mean-pairwise-correlation
formulation in Section 2.3. Across five configurations at \(b=200\),
numerical standard deviations are within 0.15\% of the first-order
prediction. Matched cluster-count distributions can still differ in
their third cumulants by 9--17\%. Thus agreement in dispersion does not
establish agreement in higher-order behaviour.

\subsection{Gaussian mean-drift
calculations}\label{gaussian-mean-drift-calculations}

For a specified independent-cluster model, the exact identity

\[\mathbb E C=\int_0^1\mathbb P\{N_b(t)\le k-1\}\,dt\]

permits deterministic convolution and numerical integration. It removes
Monte Carlo error, but quadrature, truncation, and floating-point errors
remain.

For \(m=4\), \(p_0=0.9\), and fixed Gaussian correlation \(r=0.7881\),
the recorded results are:

{\def\LTcaptype{none} 
\begin{longtable}[]{@{}
  >{\raggedright\arraybackslash}p{(\linewidth - 6\tabcolsep) * \real{0.2500}}
  >{\raggedright\arraybackslash}p{(\linewidth - 6\tabcolsep) * \real{0.2500}}
  >{\raggedright\arraybackslash}p{(\linewidth - 6\tabcolsep) * \real{0.2500}}
  >{\raggedright\arraybackslash}p{(\linewidth - 6\tabcolsep) * \real{0.2500}}@{}}
\toprule\noalign{}
\begin{minipage}[b]{\linewidth}\raggedright
\(b\)
\end{minipage} & \begin{minipage}[b]{\linewidth}\raggedright
Mean drift \(\mathbb E C-p_n\)
\end{minipage} & \begin{minipage}[b]{\linewidth}\raggedright
\(b\) times drift
\end{minipage} & \begin{minipage}[b]{\linewidth}\raggedright
Candidate smooth coefficient, multiplied by \(b/n\)
\end{minipage} \\
\midrule\noalign{}
\endhead
\bottomrule\noalign{}
\endlastfoot
25 & \(-0.006793\) & \(-0.1698\) & \(-0.1724\) \\
50 & \(-0.003418\) & \(-0.1709\) & \(-0.1723\) \\
100 & \(-0.001715\) & \(-0.1715\) & \(-0.1722\) \\
200 & \(-0.000859\) & \(-0.1718\) & \(-0.1722\) \\
400 & \(-0.000430\) & \(-0.1720\) & \(-0.1722\) \\
\end{longtable}
}

The candidate is evaluated at each \(p_n\). Resolution refinement
changed the computed expectations by less than \(2.2\times10^{-12}\).
That is a convergence check, not a certified numerical error bound. Four
further Gaussian configurations show at most 0.17\% relative discrepancy
from the candidate coefficient. These comparisons concern the Gaussian
models computed; they do not prove a remainder rate, and Section 2.4
shows the coefficient fails in general. The independent continuous
control has exactly zero drift relative to \(p_n\); exchangeability
alone does not establish that coverage law.

\subsection{Unequal cluster sizes: why the average is
insufficient}\label{unequal-cluster-sizes-why-the-average-is-insufficient}

The following Gaussian simulation uses indicator correlation \(0.5\) at
level \(0.9\). Each row fixes a size profile and repeats the cutoff
calculation 60,000 times. Predictions use the actual ceiling-rank level
\(p_n\). Only the choice of cluster-size summary changes between the two
prediction columns.

{\def\LTcaptype{none} 
\begin{longtable}[]{@{}
  >{\raggedright\arraybackslash}p{(\linewidth - 10\tabcolsep) * \real{0.1667}}
  >{\raggedright\arraybackslash}p{(\linewidth - 10\tabcolsep) * \real{0.1667}}
  >{\raggedright\arraybackslash}p{(\linewidth - 10\tabcolsep) * \real{0.1667}}
  >{\raggedright\arraybackslash}p{(\linewidth - 10\tabcolsep) * \real{0.1667}}
  >{\raggedright\arraybackslash}p{(\linewidth - 10\tabcolsep) * \real{0.1667}}
  >{\raggedright\arraybackslash}p{(\linewidth - 10\tabcolsep) * \real{0.1667}}@{}}
\toprule\noalign{}
\begin{minipage}[b]{\linewidth}\raggedright
Size profile
\end{minipage} & \begin{minipage}[b]{\linewidth}\raggedright
Mean size
\end{minipage} & \begin{minipage}[b]{\linewidth}\raggedright
Size-biased mean
\end{minipage} & \begin{minipage}[b]{\linewidth}\raggedright
Simulated sd
\end{minipage} & \begin{minipage}[b]{\linewidth}\raggedright
Predicted sd: size-biased mean
\end{minipage} & \begin{minipage}[b]{\linewidth}\raggedright
Predicted sd: plain mean
\end{minipage} \\
\midrule\noalign{}
\endhead
\bottomrule\noalign{}
\endlastfoot
Equal, size 4 & 4.00 & 4.00 & 0.0339 & 0.0335 & 0.0335 \\
Sizes 3, 4, 5 & 3.98 & 4.15 & 0.0344 & 0.0341 & 0.0336 \\
Sizes 1, 2, 4, 9 & 4.00 & 6.38 & 0.0397 & 0.0391 & 0.0322 \\
Sizes 1, 2, 4, 8, 16 & 3.76 & 8.53 & 0.0467 & 0.0466 & 0.0330 \\
\end{longtable}
}

The first two profiles have 50 clusters; the third repeats its four
sizes 13 times. The final profile has 20, 12, 8, 6, and 4 clusters of
the respective sizes. In that final row, the plain-mean prediction
understates the measured standard deviation by 29\%. This is a specified
simulation, not a measurement of generated beams. Proposition 2 explains
the size-biased weighting; it does not assert a universal 29\% error.

\subsection{PRM resampling controls and a descriptive Beta
fit}\label{prm-resampling-controls-and-a-descriptive-beta-fit}

Two checks supplement Section 5's randomized-proxy comparison. First,
retain one randomly selected prefix per question, leaving 500 singleton
clusters. Across three such pools, the average ratio of measured to
independent-Beta standard deviation is 1.002. The ratio to the same
harness's independent-row arm is 1.004. On the full clustered pool, that
harness returns a ratio of 4.44, close to Section 5's 4.46. Thus it
distinguishes the full-pool dependence from its singleton control. These
are empirical-pool resampling checks, not deployment measurements.

Second, 8,000 whole-question bootstrap repetitions compare three Beta
approximations at \(p=0.8909\). The approximation has parameters
\(p(n_e-1)\) and \((1-p)(n_e-1)\), with variance \(p(1-p)/n_e\). It is
not the exact finite-sample order-statistic Beta law.

{\def\LTcaptype{none} 
\begin{longtable}[]{@{}lll@{}}
\toprule\noalign{}
Choice of \(n_e\) & Effective size & KS distance \\
\midrule\noalign{}
\endhead
\bottomrule\noalign{}
\endlastfoot
Nominal row count & 25,028 & 0.3128 \\
Indicator-ICC plug-in & 812 & 0.0710 \\
Matched to the bootstrap variance & 1,290 & 0.0170 \\
\end{longtable}
}

The independent-row control gives distance 0.0226 against the
nominal-size approximation and inferred size 24,312. The matched
approximation describes this bootstrap distribution much better than the
other two choices. Its effective size was fitted to the same
repetitions, so this is a descriptive shape comparison. It does not
independently validate 1,290, establish tail accuracy, or provide a
goodness-of-fit test. For the raw atomic proxy, coverage takes only five
values in these repetitions. The independent-row control then has
distance 0.5054, showing why the continuous approximation fails there.

\section{Verification and
reproduction}\label{verification-and-reproduction}

The public code is available at
\url{https://github.com/ACNoonan/exceedance-design-effect} and in
archived releases. This appendix separates mathematical proofs,
numerical checks, and data-dependent interpretations.

\subsection{Mathematical verification}\label{mathematical-verification}

The proof bundle contains 59 Lean source files and a claim inventory. It
uses Lean 4.32.1 and mathlib commit
\texttt{520045ab14e26149ee970e2e617ca04b09bde5d6}. An axiom audit checks
every local declaration, including private declarations, for unapproved
axioms and proof placeholders.

Lean verifies Theorem 1, both propositions, the finite-sample guarantee,
and the variance estimator's consistency. The proofs concern the actual
pooled cutoff on the original score scale. They include the
conditional-coverage identity, ceiling ranks, initial infinite cutoffs,
and zero limiting variance. The unequal-size result uses the stated
bounded-size and mean-pair-correlation assumptions.

Both main counterexamples have checked probability constructions. The
duplicated-pair counterexample includes the iid sample, the
order-statistic expectation, and both mean limits in Section 2.4. The
zero-score-correlation example has design effect \(13/9\) at level
\(0.9\). Further checked results include the exact Beta laws, Gaussian
formulas, Bahadur representation, selection examples, and supporting
identities. The Bahadur result requires a positive derivative at the
target; the coverage limits do not require a density.

\texttt{formalization/CLAIM\_INVENTORY.md} maps these statements to
their Lean declarations and explicit hypotheses. It records no remaining
proof obligations for the listed mathematical results. The
smooth-integral calculation in Appendix B.7 is a heuristic, and Section
2.4 shows its general conclusion fails. Neither the formalization nor
the paper claims a uniform approximation-error bound. Experiments and
empirical sampling assumptions require separate validation.

Run \texttt{bash\ formalization/check.sh} from the extracted code-bundle
root. The checker may download the pinned dependencies. Run
\texttt{bash\ formalization/test\_audit.sh} to check that the audit
rejects incomplete proofs and custom axioms. The integrated compiler log
and source hashes are recorded in the accompanying verification archive.

\Needspace{18\baselineskip}

\subsection{Numerical artifacts}\label{numerical-artifacts}

{\def\LTcaptype{none} 
\begin{longtable}[]{@{}
  >{\raggedright\arraybackslash}p{(\linewidth - 2\tabcolsep) * \real{0.5000}}
  >{\raggedright\arraybackslash}p{(\linewidth - 2\tabcolsep) * \real{0.5000}}@{}}
\toprule\noalign{}
\begin{minipage}[b]{\linewidth}\raggedright
Result
\end{minipage} & \begin{minipage}[b]{\linewidth}\raggedright
Source scripts
\end{minipage} \\
\midrule\noalign{}
\endhead
\bottomrule\noalign{}
\endlastfoot
Gaussian dispersion comparison & \texttt{verify\_indicator\_icc.py},
\texttt{sim\_validation.py} \\
Score-correlation counterexample and dependence checks &
\texttt{assumption\_stress.py} \\
Tail-limit calculations & \texttt{verify\_tail\_limit.py},
\texttt{evt\_tail\_rate.py} \\
Unequal-size and ICC estimation examples &
\texttt{ragged\_and\_estimation.py}, \texttt{icc\_estimators.py} \\
Released proxy measurements and resampling &
\texttt{prm\_measurement.py}, \texttt{prm\_dispersion.py},
\texttt{test\_marginal\_scope.py} \\
SQuAD score--indicator comparison (Section 3.3) &
\texttt{score\_squad2.py}, \texttt{measure.py},
\texttt{d01\_measure.json} \\
PRM singleton control and Beta comparison (Appendix E.4) &
\texttt{deployment\_reframe.py}, \texttt{e1\_shape\_test.py},
\texttt{result\_ec01.json} \\
Document-corpus example (Section 5.3) & \texttt{score\_conll\_ner.py},
\texttt{conll\_levels.py} \\
Unweighted NHANES indicator comparison &
\texttt{one\_sided\_rho\_I.py} \\
Weighted NHANES fixed-threshold comparison &
\texttt{nhanes\_fixed\_threshold.py},
\texttt{nhanes\_samplics\_check.py} \\
Gaussian mean-drift integration & \texttt{drift\_tables.py} \\
Nested Gaussian comparison & \texttt{nested\_structure.py} \\
\end{longtable}
}

The verification archive records input hashes, software versions, and
numerical agreement for these scripts. Some analyses require cached
inputs that the archive does not include. For SQuAD,
\texttt{score\_squad2.py} creates the scores and \texttt{measure.py}
joins them to the original question contexts. The archive includes the
scoring metadata and measurement results; it does not include the cached
data or model weights. For Appendix E.4, \texttt{result\_ec01.json}
records the Beta comparison.

Deterministic convolution evaluates an exact identity numerically; it
still needs resolution and accuracy checks. Simulation summaries carry
Monte Carlo uncertainty and do not establish uniform approximation
bounds.

\section*{Acknowledgements and disclosure of AI
assistance}\label{acknowledgements-and-disclosure-of-ai-assistance}
\addcontentsline{toc}{section}{Acknowledgements and disclosure of AI
assistance}

The author thanks Ahanaf Hasan Ariq for discussion, review and feedback
during the development of this paper. His unpublished analysis of the
mean drift developed the smooth-integral calculation in Appendix B.7 and
measured its remainder against exact computation.

AI models were used throughout this work, including the research,
derivations, experiments, code, formal verification and writing. The
models were Anthropic's Claude models (Opus 5 and Fable 5) and OpenAI's
GPT-5.6 Sol.

Responsibility for every claim in this paper is the author's alone.

\bibliographystyle{plainnat}
\bibliography{references}

\end{document}